\documentclass[letterpaper, 10 pt, journal, twoside]{IEEEtran}
\usepackage{ifpdf}
\usepackage{cite}
\ifCLASSINFOpdf
  \usepackage[pdftex]{graphicx}
\else
  \usepackage[dvips]{graphicx}
\fi
\usepackage{subfigure}
\graphicspath{
	{figs/},
	{baseline_evaluation}
}

\usepackage{amsmath}
\usepackage{algorithmic}
\usepackage{array}
\ifCLASSOPTIONcompsoc
  \usepackage[caption=false,font=normalsize,labelfont=sf,textfont=sf]{subfig}
\else
  \usepackage[caption=false,font=footnotesize]{subfig}
\fi
\usepackage{dblfloatfix}
\ifCLASSOPTIONcaptionsoff
  \usepackage[nomarkers]{endfloat}
  \let\MYoriglatexcaption\caption
  \renewcommand{\caption}[2][\relax]{\MYoriglatexcaption[#2]{#2}}
\fi
\usepackage{url}
\begin{document}
\title{GFPNet: A Deep Network for Learning Shape Completion in Generic Fitted Primitives}

% \author{Michael~Shell,~\IEEEmembership{Member,~IEEE,}
%         John~Doe,~\IEEEmembership{Fellow,~OSA,}
%         and~Jane~Doe,~\IEEEmembership{Life~Fellow,~IEEE}% <-this % stops a space
% \thanks{M. Shell was with the Department
% of Electrical and Computer Engineering, Georgia Institute of Technology, Atlanta,
% GA, 30332 USA e-mail: (see http://www.michaelshell.org/contact.html).}% <-this % stops a space
% \thanks{J. Doe and J. Doe are with Anonymous University.}% <-this % stops a space
% \thanks{Manuscript received April 19, 2005; revised August 26, 2015.}}
\author{Tiberiu~Cocias,
		Alexandru~Razvant 
		and Sorin~Grigorescu% <-this % stops a space
\thanks{Manuscript received: January, 27, 2020; Revised May, 01, 2020; Accepted May, 30, 2020.}%Use only for final RAL version
\thanks{This paper was recommended for publication by Editor Cesar Cadena Lerma upon evaluation of the Associate Editor and Reviewers' comments. (Corresponding author: Tiberiu Cocias.)}%Use only for final RAL version
\thanks{The authors are with Elektrobit Automotive
		and the Robotics, Vision and Control Lab (ROVISLab),
        Transilvania University of Brasov, Romania (e-mail: 
        {tiberiu.cocias@elektrobit.com, alexandru.razvant@elektrobit.com, sorin.grigorescu@elektrobit.com}).}%
}

\markboth{IEEE Robotics and Automation Letters. Preprint Version. Accepted May, 2020}
{FirstAuthorSurname \MakeLowercase{\textit{et al.}}: ShortTitle} 




\maketitle

\begin{abstract}
In this paper, we propose an object reconstruction apparatus that uses the so-called Generic Primitives (GP) to complete shapes. A GP is a 3D point cloud depicting a generalized shape of a class of objects. To reconstruct the objects in a scene we first fit a GP onto each occluded object to obtain an initial raw structure. Secondly, we use a model-based deformation technique to fold the surface of the GP over the occluded object. The deformation model is encoded within the layers of a Deep Neural Network (DNN), coined GFPNet. The objective of the network is to transfer the particularities of the object from the scene to the raw volume represented by the GP. We show that GFPNet competes with state of the art shape completion methods by providing performance results on the ModelNet and KITTI benchmarking datasets.
\end{abstract}

% \begin{IEEEkeywords}
% IEEE, IEEEtran, journal, \LaTeX, paper, template.
% \end{IEEEkeywords}

% \fi
\IEEEpeerreviewmaketitle

\section{Introduction}

Shape completion of partial represented objects remains one of the fundamental problems in 3D perception and a key requirement for robots which interact with the physical world (e.g. mobile manipulation of objects). The main limitation is the sensor that cannot perceive the full 3D shape of the object. Therefore, many researchers focus on 3D reconstruction approaches that use one or multiple views of the object of interest to fill out the occluded information. 

The robotics and computer vision communities are tackling the problem mainly by using constraints and prior knowledge of object shapes. These solutions achieve full 3D representations by registering an a-priori volume from a database onto the perceived object \cite{Rock2015} \cite{Dame2013} \cite{Pauly2005} \cite{Sung2015}. The goal is to obtain a reconstructed volume that resembles the object's true form as close as possible. 

Deep learning paved the way to new solutions for 3D shape completion. The objective is to learn shape models by training Deep Neural Networks (DNNs) on large collections of 3D shapes. Furthermore, at runtime, the shape is used either to retrieve or to reconstruct the object of interest \cite{shapenet2015}, \cite{Varley2017}, \cite{Stutz2018CVPR}, \cite{3DEPN}, \cite{PCN_2018}. 

\begin{figure}[t]
	\centering
	\begin{center}
		\includegraphics[scale=0.5]{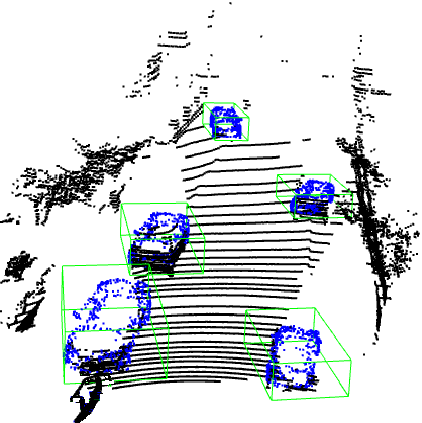}
		\caption{\textbf{Output of the GFPNet 3D shape completion apparatus}. Black points describe the input scene perceived by a LIDAR sensor, while blue points define completed 3D shapes of the objects of interest (in this particular example, traffic participants imaged in a LIDAR point cloud).}
        \label{fig:gfp_the_big_picture}
	\end{center}
	\vspace{-2em}
\end{figure}

In this paper, we focus on the specific problem of registering and modeling 3D shapes based on sparse and occluded 3D point clouds of objects, as illustrated in Fig.\ref{fig:gfp_the_big_picture}. To cope with this problem, we propose a two stage shape completion framework that \textit{i}) takes as input a point cloud, registering first a \textit{Generic Primitive} (GP) onto it, and \textit{ii}) we use GFPNet to deform (model) the appearance of the fitted GP. 

To avoid learning a specific deformation model for each type of object class, we propose to decompose the GP shape into sub-regions and generically learn a deformation model from them. The neural network behaves like a solver that has to optimize following objectives:

\begin{enumerate}
	\item re-position the GP points as close as possible to the incomplete representation of the object;
	\item ensure that the GP points remain compact and describe a smooth surface.
\end{enumerate} 

The proposed GFPNet is an improvement of our previous work \cite{cocias2013generic}, where the GP shape was modeled using a first and second order differential equations solver used to compute internal and external shape contour energies. We present here an improvement of our 3D modeling method by replacing the solver with a DNN able to outperform~\cite{cocias2013generic}, as well as other similar algorithms, especially on surfaces with strong deformations. We demonstrate the performance of GFPNet on the 3D shapes available in the ModelNet \cite{modelNet2015} and KITTI \cite{Geiger2013IJRR} datasets. 

The main contributions of the paper are: 

\begin{itemize}
  \item introduction of GFPNet as a framework for 3D shape completion from sparse and incomplete observations;
  \item a deep neural network architecture for modeling 3D surfaces in point clouds.
\end{itemize}

The rest of the paper is organized as follows. We discuss the related work in Section \ref{sec:related_work}. In Section \ref{sec:Method_overview}, we present the components of the proposed shape completion framework and give a detailed description on the DNN architecture. Subsequently, in Section \ref{sec:experimental_results}, we present the test setup used for benchmarking, along with the obtained quantitative results. Finally, conclusions are stated in Section~\ref{sec:conclusions}.

\subsection{Related Work}
\label{sec:related_work}

The problem of 3D shape completion is usually approached in robotics and computer vision from two perspectives: analytical methods \cite{Sung2015}, \cite{cocias2013generic} and deep learning algorithms for shape modeling \cite{Varley2017}, \cite{PCN_2018}. Both aim at obtaining 3D shapes from incomplete and sparse data. The analytical approaches are focused either on objects' symmetries \cite{Thrun2005}, shape retrieval \cite{Nguyen2016}, \cite{Dame2013}, \cite{Engelmann2017} or surface modeling \cite{Ahlberg1996}, \cite{Cocias_GFP_Chapter}. The main drawback of the analytical techniques is that they require a large amount of the shape's structure to be visible in the input point cloud.

Given the availability of increased computation backed by the evolution of Graphical Processing Units (GPUs), in the past years numerous deep neural network architectures were proposed for solving the 3D shape completion problem \cite{PCN_2018}, \cite{3DEPN}, \cite{Foldingnet}, \cite{3DSC_2018}, \cite{Stutz2018CVPR}, \cite{shapenet2015}, \cite{EngelmannGCPR16}, \cite{3DOR_2018}. In the following, we will address the most relevant approaches.

In \cite{PCN_2018}, the authors propose the Point Completion Network (PCN), which is a novel learning-based approach for shape completion that operates on the shape's raw point cloud without any structural assumption (e.g. symmetry) or annotation (e.g. semantic class). The method is effective across multiple object categories and works with inputs from different sensors. Unfortunately, the network fails to work on objects with disconnected parts as well as objects containing thin structures.

Acknowledging that the shape completion problem can be tackled using shape priors, in \cite{3DSC_2018} the authors proposed a weakly-supervised learning-based approach which requires no direct supervision. The network learns a shape prior on synthetic data, as well as a maximum likelihood fitting measure based on a DNN. The proposed network has difficulties completing the structure of thin objects as well as identifying the correct object category of the prior shape. This is one of the main reason why shape prior solutions are less generic. 

The shape completion process in \cite{3DEPN} uses a data-driven approach to complete partial 3D shapes through a combination of volumetric DNNs and 3D shape synthesis. The solution first infers a low-resolution, but complete output, while a 3D-Encoder-Predictor Network is used to predict and fill in the missing data. In a final pass, the authors use a patch-based 3D shape synthesis method to impose the 3D geometry from the retrieved shapes. Similar to the previous described approaches, this solution also fails to infer small and thin objects.

An important difference between GFPNet and the related work is that our network is applied on local regions instead of the entire shape at once. Given the focus of local surface modeling, the method in \cite{Ahlberg1996} uses 3D active contours to inflate/deflate raw 3D volumes depicting general sphere- or cube-like shapes. The approach has low complexity, with no prior model required to drive the deformation of the surface. The drawbacks of the concept are the input raw shapes, which have to be as similar as possible to the real shapes, as well as the convergence method, which is not guaranteed due to the global minimization of the functional energy, that may get stuck in a local minimum. One other drawback is the computational time of the algorithm, making it unsuitable for real time applications.

Approaching the modeling problem from a learning perspective, the method in \cite{Steinke2005} proposes to represent the surface of a 3D object in terms of a hyper-plane. A Support Vector Machine (SVM) is used to learn a model that deforms the surface. Although this approach implicitly solves most of the 3D active contour problems, it still has difficulties in coping with surface-holes and outliers. 

Using DNNs, the authors in \cite{Yumer2016} introduce a volumetric Convolution Neural Network (CNN) to learn deformation flows directly in the 3D Cartesian space. The network's architecture takes as input the voxelized representation of the shape, as well as a semantic deformation intention, generating a deformation flow as output. The authors claim comparable results with state of the art methods when applied to CAD models, whereas an $\approx60\%$ error rate is obtained when the algorithm is applied to single frame depth scans.

In the light of the current approaches and their limitations, we propose GFPNet for enabling shape completion of thin and complex structures from spare data. The data is represented by single point cloud observations.

\section{Method Overview}
\label{sec:Method_overview}

\begin{figure*}[t]
	\centering
	\begin{center}
		\includegraphics[scale=1]{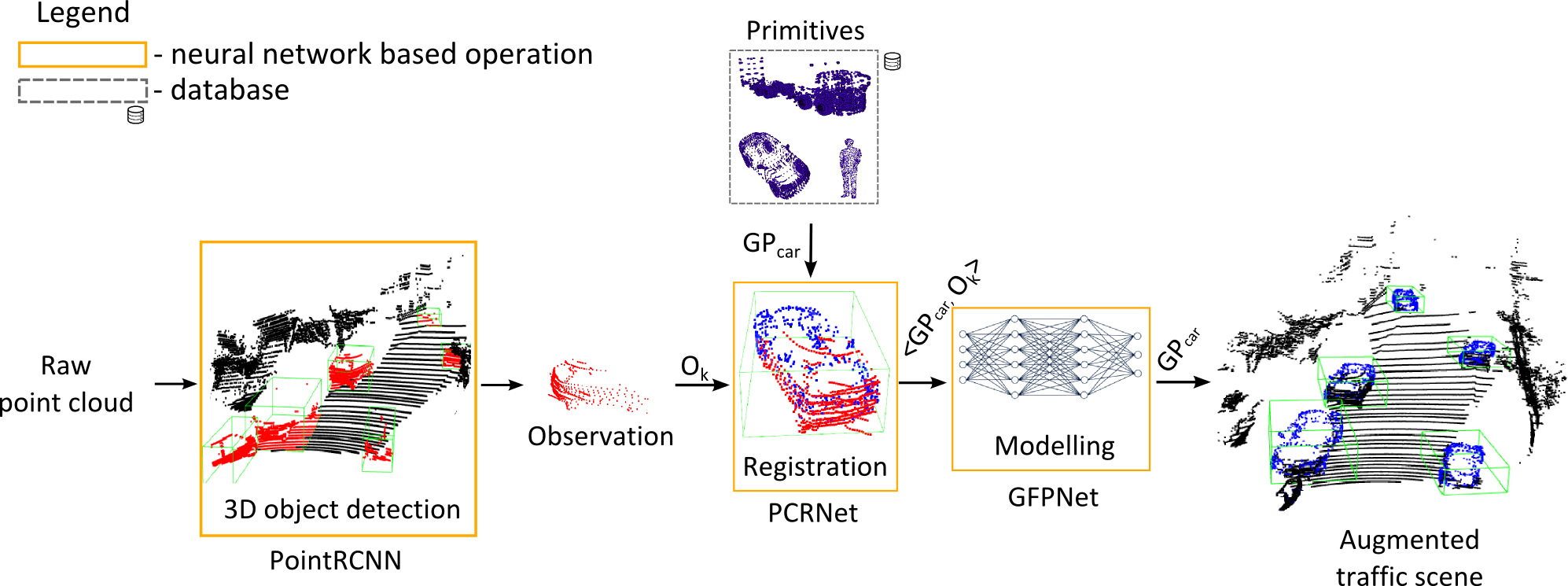}
		\caption{\textbf{Block diagram of the proposed GFPNet 3D shape completion framework}. The orange colored blocks depict DNN-based operations. We register and model a GP onto each detected object $O$, thus capturing the particular surface characteristics of $O$.}
        \label{fig:Block_diagram}
	\end{center}
	%\vspace{-2.em}
\end{figure*}

Let $O$ be a set of 3D points lying on the observed surfaces of an object that is perceived from a single perspective. Let GP be a dense set of 3D points that describe the generic shape of the observed object, while $MGP$ is a clone of the GP's point cloud whose 3D points have been re-positioned. We define the shape completion problem as predicting the $MGP$ given the GP as an initial shape and $O$ as the desired appearance (objective).

The block diagram of the proposed shape completion framework is illustrated in Fig.~\ref{fig:Block_diagram}. The proposed framework has three stages. In the first stage, we apply the PointRCNN 3D object detector \cite{PointRCNN_2019} for extracting the objects directly from the raw point cloud depicting the scene. We have chosen this particular detector based on our extensive experiments on the KITTI dataset, PointRCNN showing the best performance when compared to other state-of-the-art methods.

In the second stage, we register a GP onto the observation points in $O$. For this stage we use PCRNet's neural network \cite{PCRNet_2019}, with the objective of fining the transformation which best aligns two point clouds. As with PointRCNN, we have chosen PCRNet based on its alignment accuracy and computation time, when compared with other techniques, such as \textit{Iterative Closest Point} (ICP) \cite{BeslICP}. For computational efficiency, we limit the number of iterations to five, as the network converges rapidly. During testing, we have determined that the average run-time for registering a GP is around three milliseconds for our proposed pipeline.

Finally, in the last stage, we apply GFPNet with the purpose of modeling the surface of the GP such that it captures the particular geometries of $O$. The output is an $MGP$ that best represents the real shape that generated $O$.

%\vspace{-2em}
\subsection{Generic Primitives (GP)}
\label{subsec:gfp_definition}

A Generic Primitive (GP) is a point cloud depicting a 3D volume resembling many similar objects of the same class. For example, the GP of class \textit{car} is used to represent different types of car shapes and brands (e.g. sedan, minivan, cabrio, etc.). The GP shape is obtained using the \textit{Generalized Procrustes Analysis} \cite{Drydmard16}, by averaging the shapes of multiple objects from the same class. To ensure a complete and smooth surface of the GP we use a Moving Least Squares (MLS) surface reconstruction method to smooth and resample noisy data \cite{MLS_Smooth_2004}. We are thus ensuring that all the small errors are corrected and the so-called \textit{double walls} artefacts resulted from registering multiple shapes together are smoothed. The result is a fine shape represented by the lowest possible number of 3D points. 

\begin{figure}[b]
	\centering
		\subfigure[]{\includegraphics[scale=0.1]{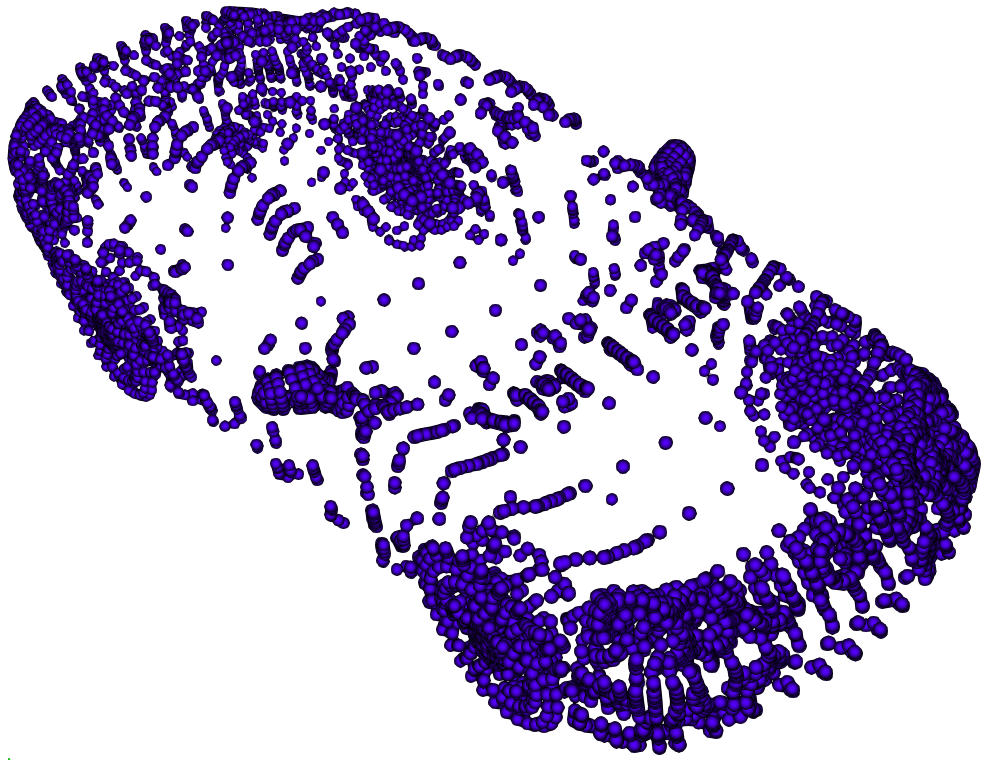}}
		\label{fig:gp_car}
		\subfigure[]{\includegraphics[scale=0.1]{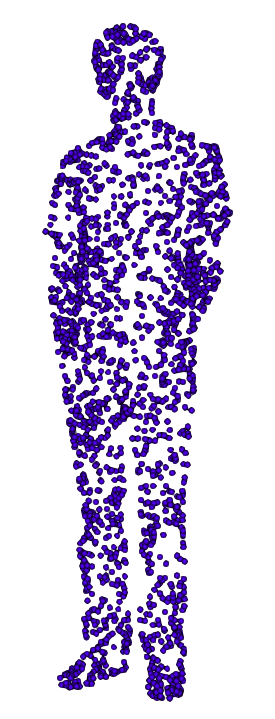}}
		\label{fig:gp_human}
		\subfigure[]{\includegraphics[scale=0.16]{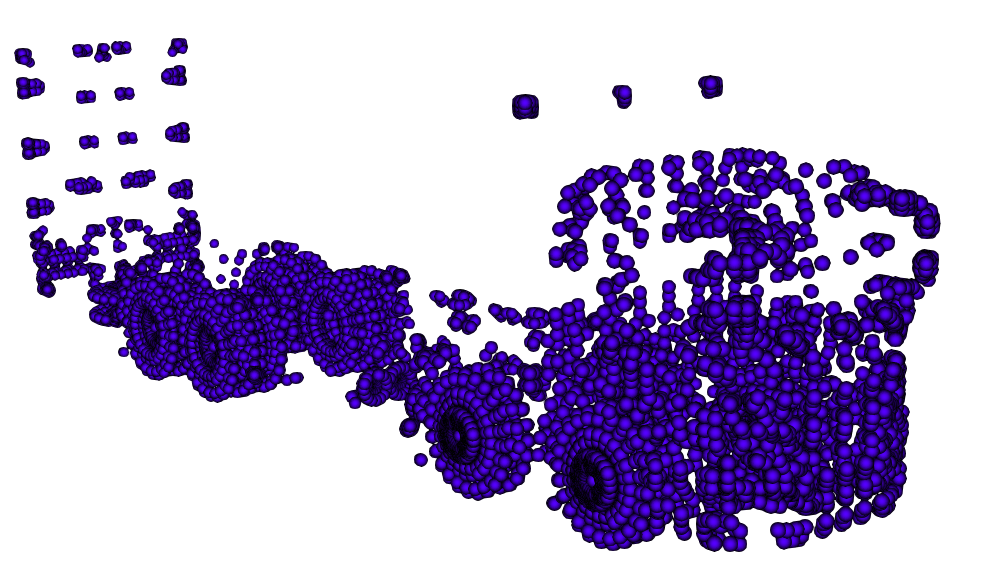}}
		\label{fig:gp_truck}	
	\caption{\textbf{3D Point clouds of various generic primitives.} (a) Car. (b) Person. (c) Truck.}
	\label{fig:generic_primitives}
	\vspace{1.em}
\end{figure}

In total, we have defined 30 GPs depicting common household items (cups, bottles, plates, etc.), as well as dynamic traffic objects (cars, pedestrians, etc.). To compute these GPs, we have used the 3D shapes from the ModelNet \cite{modelNet2015} and ShapeNet \cite{shapenet2015} datasets. Examples of car, person and truck GPs are illustrated in Fig.~\ref{fig:generic_primitives}.

%\vspace{2.em}

\vspace{1.em}
\subsection{GFPNet}
\label{subsec:GFPNet}

\begin{figure*}
	\centering
	\begin{center}
		\includegraphics[scale=0.58]{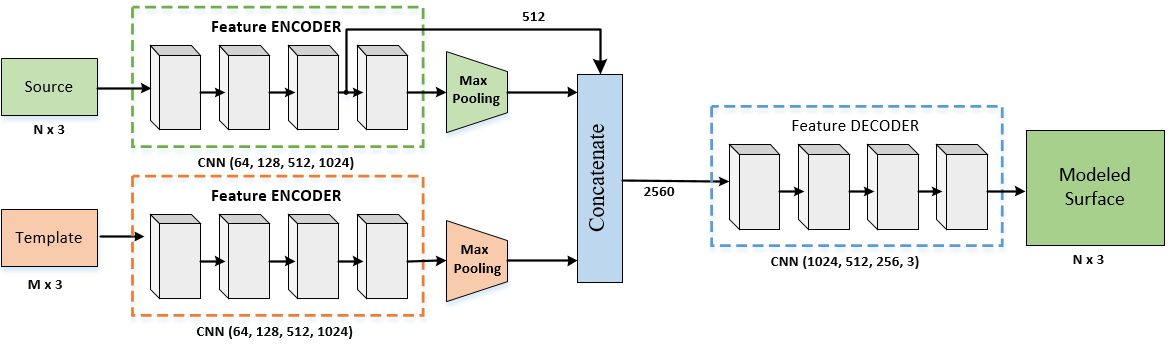}
		\caption{\textbf{GFPNet architecture}. The architecture is primarily composed of convolutional layers. $M$ and $N$ are the number of points in the template and source clouds, respectively.}
        \label{fig:GFPNet_arch}
	\end{center}
	%\vspace{-1.em}
\end{figure*}

\begin{figure}[htp]
	\centering
	\begin{center}
		\includegraphics[scale=1.0]{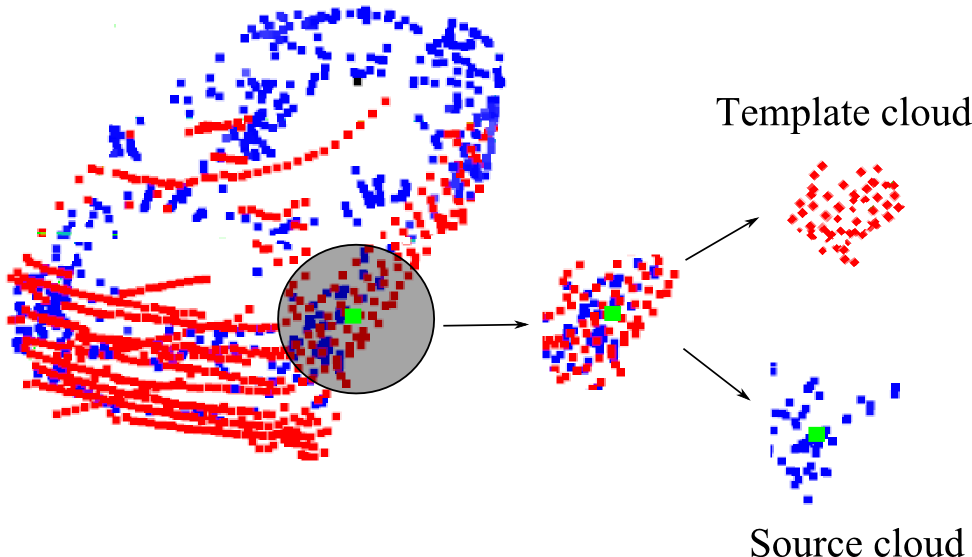}
		\caption{\textbf{Example of an input sample}. Left most is a registered GP of a vehicle (blue) aligned with the respective observation $O$ (red). The green point from the center of the gray hashed region represents point $p_i$, around which the GP surface is modeled. The right most regions represent the source $S$ (blue) and template $T$ (red) clouds.}
        \label{fig:data_preparation}
	\end{center}
	\vspace{-1.em}
\end{figure}

Our proposed DNN architecture for modelling 3D surfaces is presented in Fig.~\ref{fig:GFPNet_arch}. Given the GP registered onto $O$, let $p_i$ be the $i$-th point of the GP and $S$ a set of 3D points depicting the neighboring points of the GP that lays inside a sphere centered on $p_i$ (see Fig. \ref{fig:data_preparation}). Let $T$ be a second set of 3D points depicting the neighboring points from $O$ which lay in the same spherical area. We define $MS$ as a set of 3D points depicting a modeled version of $S$. The aim of the neural network is to calculate $MS$ by predicting the 3D point positions of $S$ given template $T$. GFPNet thus acts as a bi-objective optimizer governed by a shape representation encoded within the layers of the DNN. 

The modeling of the entire GP is achieved by applying the GFPNet modeling approach on each GP point. A relevant situation is when point $p_i$ does not have any neighboring points belonging to $O$. In this case, since there is no available template $T$, the modeled surface $MS$ will be the same as $S$.

To achieve the modeling task, the GFPNet architecture uses an encoder-decoder schema, composed of sequences of convolutional (CNN) network layers. The first half of the network behaves as a feature extractor encoding the geometrical particularities of the two inputs ($S$ and $T$), while the second half behaves as a decoder which regresses towards a modeled version of the source cloud $MS$.

The encoding of the features is performed separately for the source and template inputs, as shown in Fig. \ref{fig:GFPNet_arch}, such that the GFPNet will be able to learn a deformation model only for the source points. Each branch extracts features using CNN layers of sizes $64$, $128$, $512$ and $1024$, respectively. On each branch, we apply a symmetric max-pooling function to extract a strong feature vector. This step will ensure invariance to input permutations. Finally, the two feature vectors are glued together using a concatenation operation.

The decoder consist of a mirrored CNN, having layers of size $1024$, $512$, $256$ and $N$, where the last dimension $N$ is the number of 3D points in the input source cloud. The sizes of the CNNs were chosen based on performance tests. 

\subsection{Iterative GFP}

\begin{figure}
	\centering
	\begin{center}
		\includegraphics[scale=1.0]{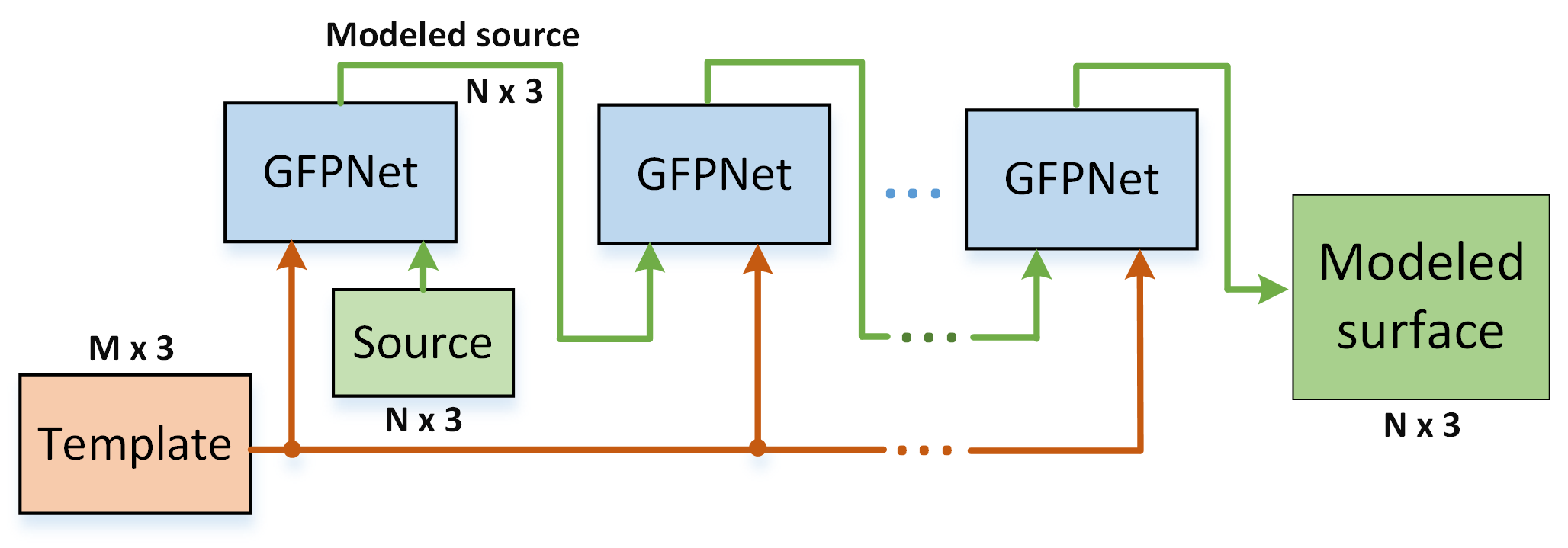}
		\caption{\textbf{Iterative GFPNet architecture used for calculating refined modeled surfaces}.  The template cloud remains the same, while the modeled source cloud is iteratively looped back into the network as a source cloud.}
        \label{fig:iterative_GFPNet_arch}
	\end{center}
	\vspace{-1.5em}
\end{figure}

Depending on the complexity of the shape that must be completed, it may be the case that a single modeling iteration will not be sufficient to converge towards the optimal shape. For this reason, we introduce the iterative scheme from Fig. \ref{fig:iterative_GFPNet_arch}, where we iteratively apply the modelling process onto the local region around each GP point $p_i$.

For the iterative implementation, the CNNs use only three hidden layers of size $64$, $128$, $1024$. In order to avoid over-fitting, we have added an additional dropout layer before the output. The reason for introducing these iterations is that it allows us to use a lower number of hidden layers, thus increasing the processing time.

In the first iteration, the template and original source clouds are fed into the GFPNet, providing a modeled source cloud as output. In the next iterations, the modeled source cloud and the template cloud are fed iteratively to the GFPNet. The process is repeated on the same point of the GP for $m$ iterations. A final primitive shape that best reproduces the observed object is obtained after modeling all GP points.

\subsection{Loss Function}
\label{subsec:loss_func}

Throughout the deformation process GFPNet optimizes two loss functions: 

\begin{enumerate}
	\item $loss_1$: minimize the distance between two point cloud densities;
	\item $loss_2$: ensure a smooth modeled surface. 
\end{enumerate}

The two objectives are combined in the following weighted loss function:

\begin{equation}
	\label{eq:multiple_losses_raw}
	loss = \alpha \cdot loss_1 + (1 - \alpha) \cdot loss_2,
\end{equation}

\noindent where, $\alpha$ is the weight factor. $loss_1$ is based on the Chamfer Distance $CD$, used to calculate the average closest point distance between the modeled GP and the observation object $O$. The loss function $loss_1$ is used to quantify how similar two surfaces are. We have chosen this particular distance metric for three reasons: \textit{i}) it solves the optimal bipartite matching problem, \textit{ii}) it is permutation invariant, while \textit{iii}) two compared point clouds do not need to be of the same size. Within GFPNet, CD is defined as:

\begin{equation}
\label{eq:cd_loss_func}
	\centering
		CD(S, T) = \frac{1}{\left | S \right |} \sum_{x \in S} \underset{y\in T}{min} \left \| x - y \right \|_2 + 
\frac{1}{\left | T \right |} \sum_{x \in T} \underset{y\in S}{min} \left \| y - x \right \|_2,
\end{equation}

\noindent where, $S$ and $T$ are the source and template point clouds, along with their respective 3D points defined by $x$ and $y$.

The smoothness of a modeled GP surface is quantified in $loss_2$ using the Laplacian operator applied to $S$. The indicator $LP(S_{i})$ gives us a measure of the surface's smoothness around point $i$ in the source cloud $S$:

\begin{equation}
	\label{eq:laplacian}
	LP(S_{i})=\frac{1}{N} \sum_{j=1}^{N} S_{j},
\end{equation}

\noindent where, $N$ is the number of neighboring points around $i$ and $S_{j}$ is the position of the $j$-th neighbor. 

We use a value of $0.7$ for $\alpha$ in Eq. \ref{eq:multiple_losses_raw}, since we are more interested in calculating shapes similar to the template.

\subsection{Dataset preparation}
\label{subsec:the_datasets}

To train and test our model, we use synthetic CAD shapes from the ModelNet database \cite{modelNet2015} and real $2.5D$ shapes from the KITTI database \cite{Geiger2013IJRR}. In the following, we detail the shape classes, statistics and the splitting of the data into training and testing sets.

\textbf{ModelNet:} we consider $939$ shapes of $4$ object categories: car ($297$), pedestrian ($108$), cup ($99$) and bottle ($435$). From this pool of shapes, we use $699$ ($75\%$) for training and $240$ ($25\%$) for testing. In order to avoid overfitting, all point clouds are augmented with an additive Gaussian noise having variance $0.1m$ and $0.01m$ for large (cars, pedestrians) and small objects (cups, bottles), respectively. 

We consider the $240$ testing CAD shapes as ground truth for evaluating the performance of the shape completion apparatus. To produce incomplete shapes needed for the testing procedure we generate partial point clouds from the ground truth shapes. For each shape, we chose 4 randomly distributed view points for generating a depth image. We then back-project these images in 3D to obtain $4$ incomplete 3D representation of the initial shape \cite{Displets2015}. Due to the fact that this strategy produces point clouds closer to real-world sensor data, it is by far more efficient than using subsets of points from the complete shapes.

\textbf{KITTI:} we have extracted shapes from KITTI's Velodyne point clouds using the provided ground truth 3D bounding boxes. We thus avoid taking into consideration points from nearby objects, such as the street, walls, or vegetation. In total, $500$ incomplete shapes depicting $2$ categories were collected ($386$ for cars and $114$ for pedestrians). We then split the shapes into training and testing sets using the same schema as for the ModelNet database.

The shapes extracted from the KITTI dataset have missing regions caused by the fact that the laser sensor is not able to image occluded surfaces. This limitation makes incomplete shapes unsuitable for usage as full ground truth data in testing. In \cite{3DSC_2018}, the authors propose a technique to generate partial ground truth shapes that can be used for evaluation purposes. Based on \cite{3DSC_2018}, we accumulate 3D point clouds of $10$ future and $10$ past frames around each object in order to reduce occlusion.
%We take into consideration the KITTI database because it enables training and testing on data that is closer to the sensory data available in real life applications.

\subsection{Training}
\label{subsec:training}

The objective of the training process is to learn a deformation model by correlating the source with the modeled cloud, where the model is encoded within the layers of the DNN from Fig. \ref{fig:GFPNet_arch}. The labels in the training set represent modeled versions of the source point cloud.

Considering the GFPNet architecture, the input data is represented by a tuple of source and template point clouds. To produce such data, we take the training shapes presented in Section \ref{subsec:the_datasets} and register a GP over them using PCRNet \cite{PCRNet_2019}. Further, for each point $i$ in the GP, we define the source point cloud as a spherical region around $i$ containing neighboring points.

From the $699$ and $375$ ModelNet and KITTI training shapes, we have extracted approx. $750k$ and $350k$ training samples, respectively. To produce labels for these samples, we use the 3D active contour modeling technique from our previous work\cite{cocias2013generic}, enhanced by applying smoothing and re-sampling via MLS on the GP points. For GP regions where the modeling process was erroneous, we manually redistribute each point using a 3D tool.

GFPNet has been trained using the Adam optimizer~\cite{Adam_optim}, while the weights have been initialized according to the scheme in~\cite{Glorot2010}. The Adam optimizer was chosen due to its adaptive learning rates. The training is based on a batch size of $256$ for $1000$ epochs, using a learning rate of $10^{-4}$ and a weight decay value of $0.92$. We have performed all training and testing operations on computing unit equipped with a single NVIDIA GeForce GTX 1080Ti GPU and an Intel Core i7 CPU, running at $4.2$ GHz.

\section{Experimental results}
\label{sec:experimental_results}

\subsection{Evaluation Metrics}
\label{subsec:evaluation}

We evaluate the GFPNet's performance on the ModelNet test set using the Chamfer Distance (CD). This distance provides a quantitative measure of similarity between the modeled GP and the ground truth shape defined in Section \ref{subsec:the_datasets}. The similarity is determined as the average closest point distance between the modeled GP and the ground truth cloud. The CD formula is the same as the one used in the loss function \ref{eq:cd_loss_func}.

Due to the fact that the CD metric can only be used to compare full shapes, as in the case of comparing modelled GFPNet shapes with the CAD models in ModelNet, we have evaluated GFPNet's performance on the KITTI test set using the approach from \cite{PCN_2018}, where the authors proposed to use the following three metrics:

\begin{itemize}
	\item \textbf{Fidelity} (F): the average distance from each point of the modeled GP to its nearest neighbor in the ground truth;
	\item \textbf{Minimal Matching Distance} (MMD): the CD between the modeled GP and the ModelNet object point cloud closest to the GP's points in terms of CD;
	\item \textbf{Consistency} (C): the average CD between the modeled GPs of the same instance in consecutive frames.
\end{itemize}

\subsection{GFPNet vs GFS}

\begin{table*}[ht]
\resizebox{\textwidth}{!}{\begin{tabular}{|c|c|ccc|c|ccc|c|c|}
\hline
\multicolumn{1}{|l|}{}                        & \multicolumn{4}{c|}{Car}                                                                                                                                                                                                                          & \multicolumn{4}{c|}{Pedestrian}                                                                                                                                                                                                                   & Cup                                                    & Bottle                                                 \\ \cline{2-11} 
\multicolumn{1}{|l|}{}                        & ModelNet                                               & \multicolumn{3}{c|}{KITTI}                                                                                                                                                               & ModelNet                                               & \multicolumn{3}{c|}{KITTI}                                                                                                                                                               & ModelNet                                               & ModelNet                                               \\ \hline
Method                                        & \begin{tabular}[c]{@{}c@{}}$CD$\\ {[}m{]}\end{tabular} & \begin{tabular}[c]{@{}c@{}}$Fidelity$\\ {[}m{]}\end{tabular} & \begin{tabular}[c]{@{}c@{}}$MMD$\\ {[}m{]}\end{tabular} & \begin{tabular}[c]{@{}c@{}}$Consistency$\\ {[}m{]}\end{tabular} & \begin{tabular}[c]{@{}c@{}}$CD$\\ {[}m{]}\end{tabular} & \begin{tabular}[c]{@{}c@{}}$Fidelity$\\ {[}m{]}\end{tabular} & \begin{tabular}[c]{@{}c@{}}$MMD$\\ {[}m{]}\end{tabular} & \begin{tabular}[c]{@{}c@{}}$Consistency$\\ {[}m{]}\end{tabular} & \begin{tabular}[c]{@{}c@{}}$CD$\\ {[}m{]}\end{tabular} & \begin{tabular}[c]{@{}c@{}}$CD$\\ {[}m{]}\end{tabular} \\ \hline
PCRNet baseline \cite{PCRNet_2019} & 0.252                                                  & 0.311                                                        & 0.275                                                   & \textbf{0.0}                                                    & 0.387                                                  & 0.411                                                        & 0.305                                                   & \textbf{0.0}                                                    & 0.109                                                  & 0.144                                                  \\ \hline
GFS \cite{cocias2013generic} & 0.040                                                  & 0.089                                                        & 0.12                                                    & 0.043                                                           & 0.057                                                  & 0.069                                                        & 0.049                                                   & 0.048                                                           & 0.027                                                  & 0.029                                                  \\ \hline
GFPNet                                        & \textbf{0.011}                                         & \textbf{0.027}                                               & \textbf{0.035}                                          & 0.018                                                           & \textbf{0.017}                                         & \textbf{0.032}                                               & \textbf{0.02}                                           & 0.027                                                           & \textbf{0.011}                                         & \textbf{0.009}                                         \\ \hline
\end{tabular}}
\caption{Performance of GFPNet against the baseline PCRNet \cite{PCRNet_2019} and our previous work on GFS \cite{cocias2013generic}}
\label{tab:per_against_prev_work}
\end{table*}

\begin{figure}
	\centering
	\subfigure[]{\includegraphics[scale=0.33]{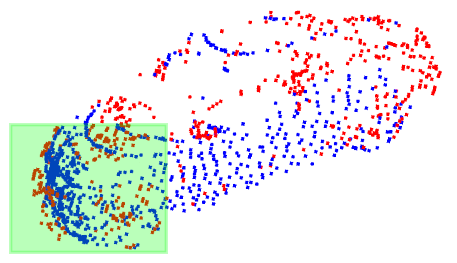}}
		\label{fig:tuple_alignement}\;\;	
		\\
		\subfigure[]{\includegraphics[scale=0.22]{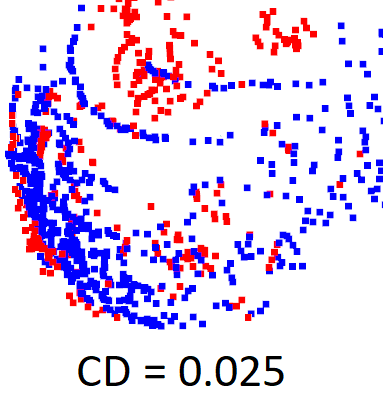}}
		\label{fig:gfpnet}\;\;
		\subfigure[]{\includegraphics[scale=0.23]{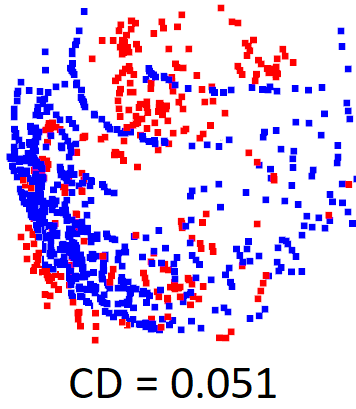}}
		\label{fig:gfs}\;\;
		\subfigure[]{\includegraphics[scale=0.18]{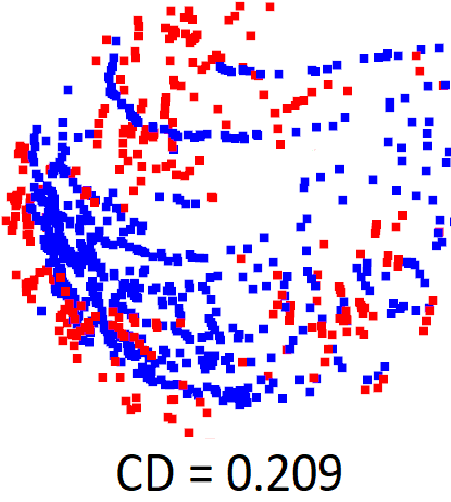}}
		\label{fig:registration_only}\;\;
	\caption{\textbf{Shape completion results over a front body part of a car object.} The computed CD for each completion approach is given bellow each picture. (a) GP registration (red) over the observation cloud $Ok$ (blue). (b) Iterative GFPNet. (c) Generic Fitted Shapes (GFS). (d) Baseline registration using PCRNet \cite{PCRNet_2019}.}
	\vspace{-1.8em}
\label{fig:reg_GFS_GFPNet}
\end{figure}

In this section, we describe the evaluation of GFPNet against our previous implementation \cite{cocias2013generic}, referred here as Generic Fitted Shapes (GFS). To understand better the relevance of the modeling step, we also provide as baseline the performance results obtained solely by registering a GP on each $O$ using the PCRNet registration method \cite{PCRNet_2019}. The GFPNet results summarized in Table \ref{tab:per_against_prev_work} have been obtained using our iterative approach with $5$ iterations. An illustration of shape completion results using the three considered approaches is shown in Fig. \ref{fig:reg_GFS_GFPNet} for the case of a car's frontal part. The least accurate results were obtained using the baseline registration method, which is not sufficient when the requirement is to produce a highly detailed reconstructed shape. 

As shown in Table \ref{tab:per_against_prev_work}, GFPNet outperforms the baseline registration algorithm and GFS in all object categories. It is important to note that the results were highly accurate for the cases of cups and bottles. This is because the shapes of these objects depict regular surfaces. For these surface types, the GFS solution to deform along the normal direction is sufficient to capture the particularities of observation $O$. However, this is not the case for the car and pedestrian object classes, since they contain highly deformed and irregular surfaces. The GFS method failed here because it had only one degree of freedom to deform the GP points along the surface normal direction. GFPNet succeeded in these cases because the DNN was trained with labels that were refined, smoothed, resampled and in some cases manually adjusted to capture as accurate as possible the template surface.

On the KITTI test dataset we have obtained a slightly higher value for the CD metric, compared to the ModelNet test dataset. This is because of the scattered distribution of the LiDAR points in KITTI, compared to the synthetic CAD models from ModelNet.

\subsection{Comparison to other methods}
\label{subsec:baseline}

In the following, we briefly describe the shape completion methods used as competing algorithms in our evaluation. Here, we compare our model against shape completion methods that work on objects from multiple classes with different levels of occlusions.

\begin{figure*}[ht]
	\centering
	\begin{center}
		\includegraphics[scale=0.6]{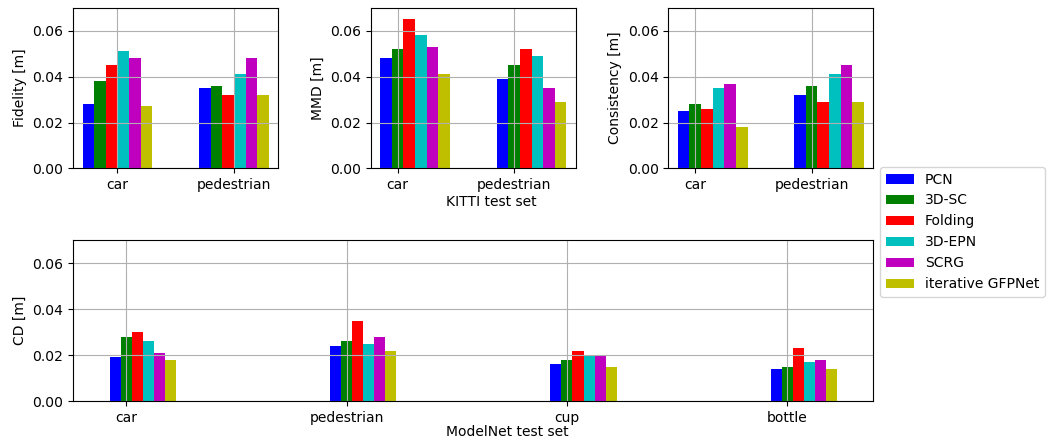}
	\end{center}
	%\vspace{-1.em}
	\caption{\textbf{Quantitative comparison on KITTI (first row) and ModelNet (second row) test datasets.} The optimal results correspond to low values for the Chamfer Distance ($CD$), Fidelity ($F$), Minimal Matching Distance ($MMD$) and Consistency ($C$).}
	\label{fig:quantitative_results_comparison}
	%\vspace{-1.em}
\end{figure*}

\begin{itemize}
	\item \textbf{PCN}: Point Completion Network \cite{PCN_2018}, which is a network that uses a raw object point cloud as input and reconstructs its shape via an encoder-decoder DNN architecture. One advantage over GFPNet is that it does not require any structural or semantic a-priori knowledge about the shape of the object.
	
	\item \textbf{3D-SC}: 3D Shape Completion under weak supervision \cite{3DSC_2018}, where a DNN learns a shape prior on synthetic data together with a maximum likelihood fitting objective. Similar to GFPNet, it uses prior knowledge about the semantics of the shape.
	
	\item \textbf{Folding} \cite{Foldingnet}, which is a network that also uses a similar encoder as GFPNet, but the decoder is purely folding-based, deforming a $128 \times 128$ 2D grid into a 3D point cloud.
	
	\item \textbf{3D-EPN}: 3D Encoder-Predictor Network \cite{3DEPN}, representing a data-driven approach to complete partial 3D shapes through a combination of volumetric DNNs and 3D shape synthesis. For comparing the outputs, we have converted the output of 3D-EPN into point clouds by extracting the isosurface around a small area and then uniformly resampling the obtained cloud.
	
	\item \textbf{SCRG}: Shape Completion enabling Robotic Grasping \cite{Varley2017}, which is a convolution neural network trained to complete an object's mesh representation. It does not use any prior structural and semantic information. In order to calculate $CD$, we have converted the resulted voxel into point clouds. 	
\end{itemize}

For all the above approaches, we have used already-trained models from the authors, since their extensive experiments were conducted on the same or similar datasets as ours (KITTI\cite{Geiger2013IJRR}, ModelNet\cite{modelNet2015} or ShapeNet\cite{shapenet2015}).

\subsection{Ablation study}

\begin{figure}[ht]
	\centering
	\begin{center}
		\includegraphics[scale=0.55]{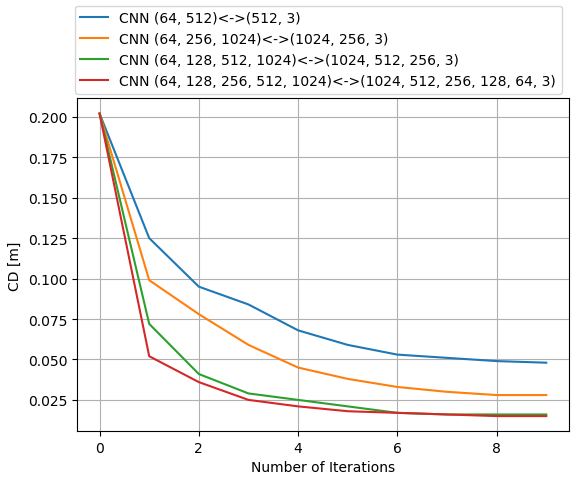}
	\end{center}
	\vspace{-1.em}
	\caption{\textbf{The value of $CD$ for different numbers and sizes of the CNN layers in the GFPNet architecture}. The outcome of the experiment corresponds to an encoder-decoder DNN architecture composed of 4 convolutional layers of size $64$, $128$, $512$ and $1024$ and an optimal value of $5$ iterations.}
	\label{fig:ablation_iteration_vs_mlp}
		\vspace{-1.em}
\end{figure}

The goal of the ablation study is to show the importance of the different components in our architecture. As illustrated in Fig. \ref{fig:ablation_iteration_vs_mlp}, we have varied the structures of the CNN layers in the encoder and decoder network, as well as the number of iterations. 

For this study, the sizes and number of convolutional layers in the GFPNet's encoder and decoder architecture from Fig.~\ref{fig:GFPNet_arch} have been varied. We have observed that using only two convolutional layers affects the quality of the generated feature vector, determining the overall modeled surface to be more rigid and thus unable to fold over deformed surfaces. On the other hand, using more than four convolutional layers produces overfitting and increased run-time.

To identify the optimal number of convolution layers, we have applied the evolutionary approach described \cite{NeuroTraj}. Aiming to obtain a balance between run-time and precision, we have identified a DNN architectures composed of 4 convolutional layers of size $64$, $128$, $512$ and $1024$ for the encoder and decoder, respectively.

In order to determine the optimal number of iterations, we have measured the evolution of $CD$ over an evaluation subset of shapes. The optimal value of 5 iterations was empirically determined.

\vspace{-1em}
\subsection{Discussion}

The performance of the competing shape completion methods is shown in Fig. \ref{fig:quantitative_results_comparison}, taking into account the evaluation metrics presented in Section \ref{subsec:evaluation}. The iterative GFPNet, configured to perform 5 iterations, provided the better results on the majority of shapes from both test sets.

GFPNet obtained similar results to PCN for the cup, bottle and car test shapes, while outperforming it by a considerable margin for the pedestrian shapes. This is because the PCN Folding based multistage decoder considers weak constraints on the local densities. On inputs where the object structure has an occlusion ratio above $60\%$ (e.g. a vehicle seen only from the back), the coarse PCN output fails to produce a correct global shape.

On the other hand, the Folding approach often produces outliers that are not consistent with the global shape. This reflects in high $CD$, $F$ and $MMD$ values for all compared methods. Similar to GFPNet, the Folding network can be applied also on local surfaces. Nevertheless, in order to achieve an accuracy similar to GFPNet, it requires a considerable higher number of iterations.

The SCRG network has a good performance on regular shapes, such as cups and bottles, but fails on cars and pedestrians. Especially in the completed part, the network produces a very rough approximation mainly due to the lack of prior information on the object class. The same behavior has also been obtained on the 3D-EPN volumetric approach. Also in this case, GFPNet performs better in both the $CD$, $F$ and $MMD$ measures. 

\section{Conclusions}
\label{sec:conclusions}

In this paper, we have introduced GFPNet, which is a 3D volumetric object modeling approach for objects described by incomplete point clouds. Its main goal is to deliver precise 3D shape models and pose estimation for objects present in different real-world scenes. The main novelty of the algorithm is represented by the usage of a deep learning technique for deforming a Generic Primitive (GP) with the purpose of 3D shape completion.

As future work, we plan to apply GFPNet for modeling a larger category of object shapes, as well as to non-point cloud data such as images, where the shape completion operation should be performed solely on 2D visual information.

\section*{Acknowledgment}

The authors would like to thank Elektrobit Automotive for the infrastructure and research support.

\bibliographystyle{IEEEtran}
\bibliography{references}


\vspace{-3em}
\begin{IEEEbiography}[{\includegraphics[width=1in,height=1.25in,clip,keepaspectratio]{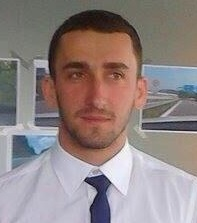}}]{Tiberiu Cocias} received the Ph.D. degree in System Engineering in 2013 and the Dipl.-Eng. degree in Control Engineering and Computer Science in 2009 both from Transilvania University of Brasov, Romania. He is a member of the ROVIS team, as well as team manager in the Artificial Intelligence Tech Centre at Elektrobit Automotive Romania. His main areas of research are 3D perception, 3D object reconstruction, object segmentation, volumetric shape modelling and object volume approximation. Since 2015 he is a lecturer at the Department of Automation, Transilvania University of Brasov, Romania, where he teaches Robotics, Expert Systems and Computer Programming.
\end{IEEEbiography}

\vspace{-2em}
\begin{IEEEbiography}[{\includegraphics[width=1in,height=1.25in,clip,keepaspectratio]{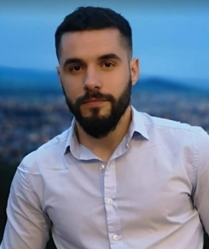}}]{Alexandru Razvant} received his Dipl.-Eng. degree in Information Technology in 2019 from Transilvania University of Brasov, Romania. He is also affiliated with the Artificial Intelligence Tech Centre at Elektrobit Automotive Romania, where he is a software developer. His main areas of research are 3D perception, 3D object reconstruction, object segmentation, volumetric shape modelling and object volume approximation.
\end{IEEEbiography}

%\newpage

\vspace{-2em}
\begin{IEEEbiography}[{\includegraphics[width=1in,height=1.25in,clip,keepaspectratio]{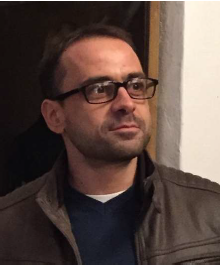}}]{Sorin Grigorescu}
received the Ph.D. degree in Robotics from the University of Bremen, Germany, in 2010 and the Dipl.-Eng. degree in Control Engineering and Computer Science from Transilvania University of Brasov, Romania, in 2006. Sorin is an associate professor at the Department of Automation, Transilvania University of Brasov, Romania. Since June 2013, he is also affiliated with Elektrobit Automotive, where he is the global Head of Artificial Intelligence (AI), coordinating an international team of AI researchers working on deep learning solutions for the automotive industry. His research interests are Computer Vision, Artificial Intelligence for Autonomous Vehicles, Learning Controllers, Rehabilitation Robotics and Feedback Control in Computer Vision.
\end{IEEEbiography}


%\vfill


\end{document}